\documentclass[journal]{IEEEtran}
\usepackage{graphicx}
\usepackage{tabularx}
\usepackage[edges]{forest}
\usepackage{enumitem} 
\usepackage{longtable}
\usepackage{array}
\newcolumntype{L}[1]{>{\raggedright\arraybackslash}p{#1}}  
\usepackage{minted}
\setminted{fontsize=\footnotesize, breaklines, autogobble}

\usepackage[ruled,vlined]{algorithm2e}
\usepackage{amsmath,amssymb}
\usepackage{booktabs}

\usepackage[caption=false,font=footnotesize]{subfig}

\usepackage[T1]{fontenc}
\usepackage[utf8]{inputenc}

\DeclareUnicodeCharacter{1E24}{H}    
\DeclareUnicodeCharacter{02BB}{'}    
\DeclareUnicodeCharacter{0100}{\=A}  
\DeclareUnicodeCharacter{0101}{\=a}  
\DeclareUnicodeCharacter{2011}{-}    
\DeclareUnicodeCharacter{2013}{--}   
\DeclareUnicodeCharacter{2014}{---}  
\usepackage{hyperref}

\begin{document}

\title{An Outcome-Based Educational Recommender System}

\author{Nursultan Askarbekuly,
        Timur Fayzrakhmanov,
        Sladjan Babarogić,
        Ivan Luković.
\thanks{N. Askarbekuly is with Faculty of Computer Science and Engineering, Innopolis University, Innopolis,
Russia and with the Faculty of Organizational Sciences, The University of Belgrade, Belgrade,
Serbia, e-mail: n.askarbekuly@innopolis.university}
\thanks{T. Fayzrakhmanov  is with Faculty of Computer Science and Engineering, Innopolis University, Innopolis,
Russia e-mail: t.fayzrakhmanov@innopolis.university}
\thanks{Sladjan Babarogić is with the Faculty of Organizational Sciences, The University of Belgrade, Belgrade, Serbia e-mail: sladjan.babarogic@fon.bg.ac.rs}
\thanks{Ivan Luković is with the Faculty of Organizational Sciences, The University of Belgrade, Belgrade, Serbia e-mail: ivan.lukovic@fon.bg.ac.rs}
\thanks{Manuscript received Month date, year; revised Month date, year.}}

\markboth{Journal X}%
{Askarbekuly et al: An Outcome-based Educational Recommender System}

\maketitle

\begin{abstract}
Most educational recommender systems are tuned and judged on click- or rating-based relevance, leaving their true pedagogical impact unclear.  
We introduce \textbf{OBER}—an \emph{Outcome-Based Educational Recommender} that embeds learning outcomes and assessment items directly into the data schema, so any algorithm can be evaluated on the mastery it fosters.
OBER uses a minimalist entity-relation model, a log-driven mastery formula, and a plug-in architecture.  
Integrated into an e-learning system in non-formal domain, it was evaluated trough a two-week A/B/C test with over 5 700 learners across three methods: fixed expert trajectory, collaborative filtering (CF), and knowledge-based (KB) filtering.
CF maximized retention, but the fixed path achieved the highest mastery. 
Because OBER derives business, relevance, and learning metrics from the same logs, it lets practitioners weigh relevance and engagement against outcome mastery with no extra testing overhead.
The framework is method-agnostic and readily extensible to future adaptive or context-aware recommenders.
\end{abstract}

\begin{IEEEkeywords}
recommendation systems, e-learning, evaluation, assessment, intended learning outcomes, constructive alingment, empirical software engineering.
\end{IEEEkeywords}

\IEEEpeerreviewmaketitle

\section{Introduction}
\IEEEPARstart{E}{ducational recommender systems} \textbf{(ERS)} have seen widespread use over the past two decades \cite{rivera2018recommendation, drachsler2015panorama}. ERS primarily aim to personalize content delivery to help learners acquire the most relevant knowledge and skills \cite{urdaneta2021recommendation}, thereby optimizing the achievement of learning outcomes. 

A learning outcome defines what a learner is expected to know, understand, or be able to do as a result of a learning experience \cite{spady1994outcome, europeanQualification}. \emph{Constructive alignment} \cite{biggs2012student} is a foundational pedagogical principle that places intended outcomes at the center of the educational process. It holds that both learning activities and assessments must be deliberately aligned with the outcomes—either by contributing to their development or by measuring their attainment. In our previous work \cite{askarbekuly2021building}, we demonstrated that constructive alignment and outcome-based evaluation can be successfully integrated within educational software. We therefore argue that any recommender system aimed at promoting learning must also measure it to evaluate its effectiveness.

However, few existing systems incorporate learning outcomes directly into their learner or domain models, and even fewer attempt to evaluate the educational impact of recommendations \cite{our_slr}. Instead, ERS evaluation often focuses on engagement metrics such as click-through rates or user ratings. This misalignment between goals and evaluation may stem from the assumption that engagement correlates with learning, or from a lack of mechanisms to assess learning impact efficiently.

To address this gap, we introduce an \emph{Outcome-Based Educational Recommender} (OBER) system. OBER is method-agnostic and embeds learning outcomes and assessment items directly into its data schema. This structure allows for verifying outcome achievement, thereby enabling educationally meaningful evaluation of different recommendation strategies.

Our central hypothesis is: \emph{The proposed outcome-based educational recommender can serve as a mechanism to verify learning outcomes, allowing us to assess the effectiveness of various recommendation methods.}

By addressing this hypothesis, we offer a practical approach to evaluating personalized learning trajectories through an outcome-based lens. Automating this evaluation is key to iterating on recommendation strategies and enhancing their educational impact.

The remainder of this paper reviews related work, presents the OBER framework, explains the evaluation procedure, and concludes with a discussion of findings and directions for future research.

\section{Background and Related Work}

In this section we give a concise introduction to recommender systems, then review how they have been applied in educational settings.  We focus on the core data inputs and algorithms—following Burke’s framework \cite{burke2002hybrid}—and highlight the gap in outcome‐based evaluation for non-formal and informal learning.

\subsection{Recommender Systems Overview}

Recommender systems (RS) help users discover items of interest by predicting their preferences.  Most RS approaches draw on three fundamental data sources:

\begin{itemize}
  \item \emph{User Profile:} personal attributes of the user receiving recommendations (e.g.\ demographics, preferences).  
  \item \emph{Item Descriptions:} metadata or content features (e.g.\ keywords, categories, skills required).  
  \item \emph{Interaction Logs:} historical records of learner–item interactions (clicks, ratings, completions).  
  \item \emph{Knowledge Relations:} explicit domain information linking the items to other items or to learners (e.g.\ prerequisites, topic hierarchies).
\end{itemize}

Different recommendation algorithms combine these inputs in various ways:

\begin{enumerate}
  \item \textbf{Collaborative Filtering (CF)} \cite{burke2002hybrid}  
    Learns patterns in the \emph{interaction logs} and recommends \emph{items} that peers with similar histories enjoyed.  CF does not require detailed item metadata or domain rules.
  \item \textbf{Content‐Based Filtering (CBF)}  
    Uses the \emph{item descriptions} and \emph{the user profile} to match items whose features resemble those the user has previously liked.  CBF can avoid repeating already‐seen items by consulting the \emph{interaction logs}.
  \item \textbf{Knowledge‐Based Filtering (KB)}  
    Leverages \emph{knowledge relations}—such as prerequisite graphs—to infer which items are appropriate next.  KB methods require a domain model (often hand‐crafted) to encode the dependencies.
  \item \textbf{Hybrid Approaches}  
    Combine two or more of the above techniques to offset each method’s weaknesses and exploit complementary strengths \cite{burke2002hybrid}.
\end{enumerate}

Table~\ref{tab:ers_inputs} illustrates which inputs each method relies on most heavily:

\begin{table}[h]
  \centering
  \begin{tabular}{lcccc}
    \toprule
    \textbf{Method} & \textbf{Items Metadata} & \textbf{Profile} & \textbf{Logs} & \textbf{Knowledge} \\
    \midrule
    CF    &      –         &   –              &   +           &   –               \\
    CBF   &      +         &   +              &   +           &   –               \\
    KB    &      +         &   +              &   –           &   +               \\
    Hybrid&    +/–         &   +/–            &   +/–         &   +/–             \\
    \bottomrule
  \end{tabular}
  \caption{Core inputs for common recommendation methods.}
  \label{tab:ers_inputs}
\end{table}

Apart from these traditional methods, new approaches have also emerged: deep learning–based recommenders \cite{deeplearning2022} can  capture complex user–item patterns, and reinforcement‐learning methods \cite{reinforcement2022} adapt policies online as learner behavior evolves. Still even these methods will typically be trained on a combination of the same data elements.

\subsection{Data Modeling in Educational Recommender Systems}

Educational recommender systems (ERS) use the same data entities: users, items, interaction logs, and knowledge relations. The users are usually learners, while the recommended items are mostly learning resources, and the logs are the record of learner-item interactions \cite{our_slr}.

In addition to that, research on ERS suggests to enrich learner profile with:
\begin{itemize}
    \item prior knowledge \cite{drachsler2015panorama}
    \item learning goals \cite{fuzzyTree7094243} \item learning style \cite{blended7104183, ontology10, ersdoDynamicOntology}
    \item language proficiency \cite{SocialRetrieval2018}.
\end{itemize}

Knowledge relations are also common in ERS. \cite{fuzzyTree7094243} notes that usually there exist precedence relations between items. In other words, certain activities need to be completed before others to ensure a coherent and effective learning experience. Therefore, they stress the importance of having a graph that shows relations between nodes. To establish the relations, various techniques have been suggested, including manual specification or automatic methods such as association rules \cite{largescale}.

The key difference of ERS from other recommenders is the end-goal of maximizing learning in terms of acquired knowledge, skills, and competencies. This basically means maximizing the learning outcomes.

\subsection{Learning Outcomes and Assessment in ERS}
Outcome-based approach \cite{spady1994outcome} became the de-facto standard for learning and teaching \cite{europeanQualification}. It has been widely acknowledged that aligning assessments with intended learning outcomes enhances the effectiveness of the educational process \cite{biggs2012student}.

\emph{Learning outcomes} are specific statements of what the learners are expected to achieve. \emph{Outcomes} must be measured and verified, and that is the function of \emph{assessment} \cite{elearning_keys}. A slightly different term is \emph{evaluation}, which measures the effectiveness of an educational program or product. Although similar, the assessment has the learner as its object, while evaluation focuses on the methods and tools. Using these definitions, we can state the goal of our outcome-based educational recommender (OBER) in the following way: enable evaluation of educational recommender systems by verifying outcomes through assessment.

In one of our earlier works, we have demonstrated that it is possible to align assessment activities with the intended outcomes when building an educational product  \cite{askarbekuly2021building}. In fact, the outcome-based approach is similar to the goal-modeling in requirements engineering, i.e. the high level business goals are refined into lower-level objectives and then connected to the corresponding functionality and metrics \cite{askarbekuly2020combining}. Therefore, it should be possible to derive assessment activities from the intended outcomes and to use them for evaluation.

However, assessment results are rarely used for evaluation in educational recommender systems (ERS) \cite{our_slr}. 
Instead, the most common metric is relevance, typically derived from user ratings. While optimizing for user ratings may be effective in entertainment contexts, in education this is analogous to recommending resources solely based on previously liked material, without measuring their actual impact on learning outcomes. Fazeli et al. \cite{usercentric7994718} show that this rating-based approach is insufficient and advocate for supplementing it with additional evaluation methods in future work.

To the best of our knowledge, no ERS to date has incorporated outcome assessment in non-formal or informal educational settings \cite{our_slr}. The assessment of learning outcomes has so far been conducted only in the context of formal university courses \cite{blended7104183, scratch8651403}, where standardized tests are readily available. 

To summarize, we need to measure the mastery of learning outcome to evaluate the effectiveness of an educational recommendation system (ERS). The learning outcomes are usually measured through assessment. Existing research on ERS appears to overlook this major aspect, particularly in the non-formal and informal educational domains. To fill this gap, we propose an outcome-based data schema in the following section.


\section{OBER: Outcome-Based Educational Recommender}

In this section, we present the key elements of the proposed OBER system,  which is the abbreviation for Outcome-Based Educational Recommender. The main goal is to facilitate the mastery and assessment of learning outcomes. We demonstrate a step-by-step method to integrate it with an e-learning system. As the case study project, we collaborated with NamazApp, a popular mobile e-learning application that helps beginners to learn Muslim prayer. Being an e-learning system from the non-formal educational domain, NamazApp was a good fit to test our main hypothesis.

\begin{figure}[h]
    \centering
    \includegraphics[width=0.45\textwidth]{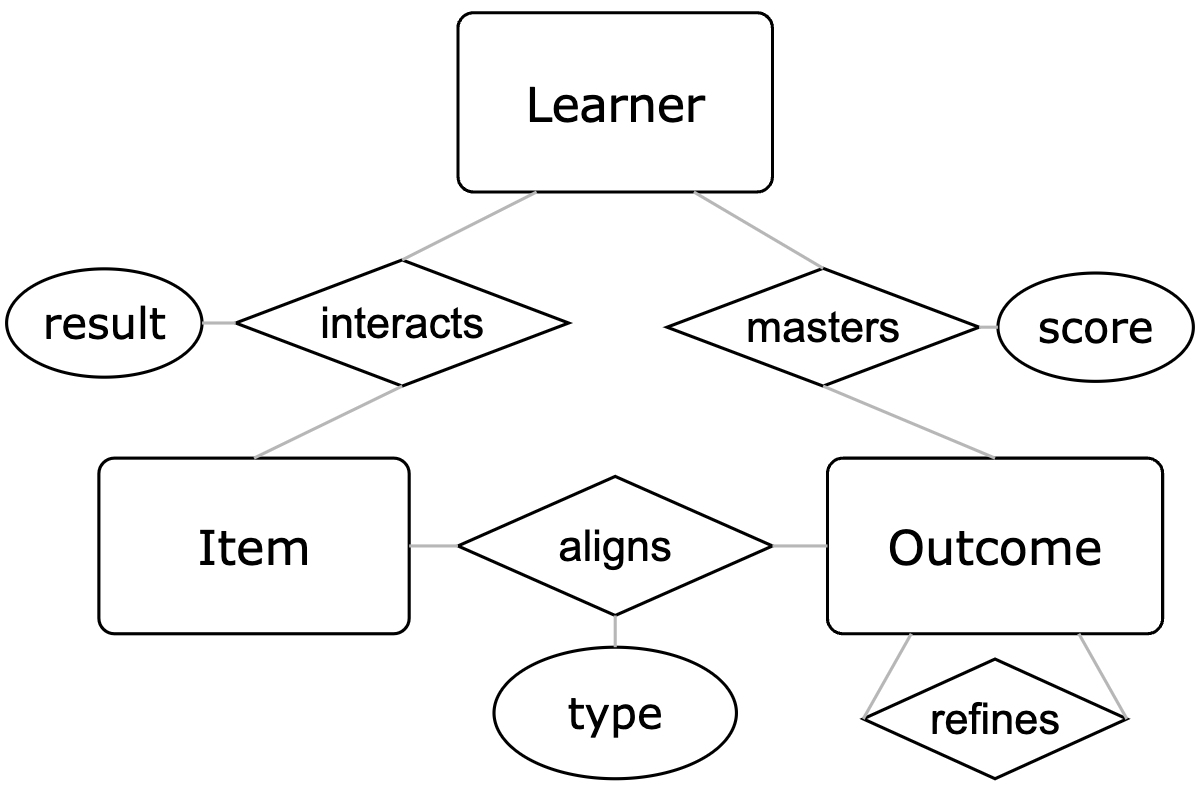}
    \caption{The data schema}
    \label{fig:data_schema}
\end{figure}

Figure~\ref{fig:data_schema} shows the conceptual Entity–Relationship schema for tracking learner-item interactions and the hierarchical structure of learning outcomes.  

The core entities (rectangles) in Fig.~\ref{fig:data_schema} are:

\begin{itemize}
  \item \textbf{Learner:} the end-users who consume and are recommended learning items.
  \item \textbf{Item:} any piece of content or activity in the system (articles, quizzes, exercises, etc.)
  \item \textbf{Outcome:} a competency or learning goal that the system tracks and promotes.
\end{itemize}

The \emph{Outcome} entity is the main addition to the standard \emph{Learner-Item} combination common in ERS. This addition changes the system's from being engagement-oriented to becoming outcome-oriented. It enables the evaluation of learning outcomes within any given recommendation process.

Each relationship (diamond) now carries its own attribute:

\begin{itemize}
  \item \textbf{interacts} (between \emph{Learner} and \emph{Item}): captures each time a learner engages with an item, with a \textit{result} attribute indicating the outcome of that interaction (e.g.\ pass/fail, time spent, correctness).
  \item \textbf{masters} (between \emph{Learner} and \emph{Outcome}): reflects the learner’s demonstrated level of mastery of an outcome, with a \textit{score} attribute (e.g.\ a percentage or proficiency level).
  \item \textbf{aligns} (between \emph{Item} and \emph{Outcome}): defines how each item is linked to an outcome, with a \textit{type} attribute specifying whether the item “promotes” (contributes) or “verifies” (assesses) that outcome.
\end{itemize}

Finally, an \emph{Outcome} can be broken down into sub-outcomes via a recursive \textbf{refines} relationship: high-level competencies are connected to more granular skills, forming an outcome-tree (or forest).

We took a minimalist approach to the schema on purpose, to keep it extensible and agnostic to the recommendation method. In the following subsections, we demonstrate the schema in action by integrating it into NamazApp in several consecutive steps:
\begin{enumerate}
    \item Establish learning outcomes;
    \item Align outcomes with items;
    \item Define mastery formula.
\end{enumerate}

The following subsections describe each step in detail.

\subsection{Establish the learning outcomes}

Often non-formal and informal e-learning systems do not have explicitly formulated learning outcomes, this was also the case with NamazApp.  In such cases, the engineer needs to facilitate their explicit formulation by a domain expert, possibly using semi-automatic means \cite{Askarbekuly2024}.

When designing the learning outcomes, a good starting point is the highest-level goal that a given e-learning systems aims to achieve. This goal is the root of the \emph{Outcomes tree}. This goal is then refined based on the domain specifics, and it is best to reuse existing structures.

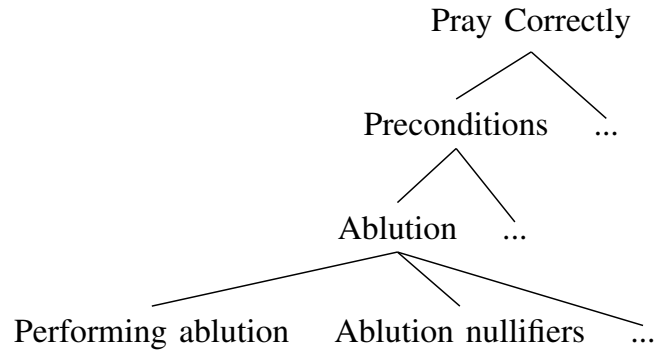
\begin{figure}[ht]
  \centering
  \resizebox{\columnwidth}{!}{%
    \begin{forest}
      for tree={
        grow=south,
        parent anchor=south,
        child anchor=north,
        edge={-},
        l sep=1em,
        s sep=0.5em,
        font=\small
      }
      [Pray Correctly
        [Preconditions
              [Ablution
                [Performing ablution]
                [Ablution nullifiers]
                [...]
              ]
              [...]
        ]
        [...]
      ]
    \end{forest}%
  }
  \caption{Snippet of the Outcomes Tree for Namaz App.}
  \label{fig:tree}
\end{figure}

NamazApp's goal is to help beginners pray in a correct way. Starting for this root, we then used the classification by the Hanafi school of law \cite{AlShurunbulali2010} to define what 'correct prayer' means in terms of lower-level knowledge and skills. A snippet of the resulting Outcomes tree is shown in Figure \ref{fig:tree}.

Listing~\ref{lst:outcomes-json} shows a data sample of how the outcome tree can be stored in JSON format.

\begin{listing}[h]
\caption{Outcomes JSON (excerpt)}\label{lst:outcomes-json}
\begin{minted}{json}
[
  {
    "id": "pray_correctly",
    "title": "Pray correctly",
    "description": "Perform the five daily prayers in accordance with the Hanafi school.",
    "parent_id": null
  },
  {
    "id": "preconditions",
    "title": "Preconditions",
    "parent_id": "pray_correctly"
  },
  {
    "id": "ablution",
    "title": "Ablution (wudu')",
    "parent_id": "ritual_purity"
  },
  {
    "id": "performing_ablution",
    "title": "Performing ablution",
    "parent_id": "ablution"
  },
  {
    "id": "ablution_nullifiers",
    "title": "Ablution nullifiers",
    "parent_id": "ablution"
  },
  ...
]
\end{minted}
\end{listing}

\subsection{Align the outcomes with learning items}
In a constructively aligned e-learning system, every learning outcome must be covered by at least one item that \emph{verifies} its achievement. 

Listing \ref{lst:items-json} shows a data sample of how a learning item can be stored in JSON format.

\begin{listing}[h]
\caption{Items JSON (excerpt)}\label{lst:items-json}
\begin{minted}{json}
[
  ...,
  {
    "id": "ablution_practice",
    "title": "Practice: performing ablution",
    "type": "exercise"
  },
  {
    "id": "ablution_obligatory",
    "title": "Obligatory actions in ablution",
    "type": "mutlipce_choice_quiz"
  },
  {
    "id": "what_nullifies_ablution",
    "title": "What nullifies ablution?",
    "type": "mutlipce_choice_quiz"
  },
  ...
]
\end{minted}
\end{listing}

Mapping the outcomes and items can be a non-trivial task if there are many items available within the system. For NamazApp, we used an LLM to produce the initial alignments. The input was the set of all outcomes and the set of all available items. The output was a mapping between the outcomes and the learning items that verify them. The mapping was further checked and edited by the NamazApp team.

Listing~\ref{lst:alignment-mappings} shows a JSON excerpt of how “Ablution” outcomes were mapped to learning items. Using these alignment mappings, we can audit coverage and see which outcomes are under-assessed. Most importantly, having the mappings, we are now ready to measure the learners' mastery of each outcome.

\begin{listing}[h]
\caption{Alignment mappings JSON (excerpt)}\label{lst:alignment-mappings}
\begin{minted}{json}
[
  {
    "outcome_id": "performing_ablution",
    "learning_item_ids": [
      "ablution_practice",
      "ablution_obligatory"
    ]
  },
  {
    "outcome_id": "ablution_nullifiers",
    "learning_item_ids": [
      "what_nullifies_ablution"
    ]
  }
]
\end{minted}
\end{listing}

\subsection{Calculate Outcome Mastery}

To calculate the mastery of outcomes for a particular learner, we use:

1. Alignment mappings $A$:
\[
A = \{(o,i) : \text{outcome }o \text{ is verified by item } i \}.
\]

2. Interaction logs $I_l$ for learner $l$:
\[
I_l = \{(i,r): \text{learner } l \text{ tried item } i \text{ and got result } r\}.
\]

3. Outcome mastery $M_l$:
\[
M_l = \{(o,s): \text{learner } l \text{ has mastery score } s \text{ for outcome } o\}.
\]

The first two sets are the input and the third set is the output of our algorithm:

\begin{algorithm}[h]\footnotesize
\SetAlgoLined
\KwIn{

$I_l$: interaction logs for learner $l$;

$A$: alignment mappings}
\KwOut{

$M_l$: mastery scores per outcome}

-----

$M_l = \{\}$\;

\ForEach{outcome $o$}{
    $best\_r = 0$\;
    \ForEach{item $i$ such that $(o,i) \in A$}{
        retrieve $r$ such that $(i,r) \in I_l$\;
        \If{$r > best\_r$}{
            $best\_r = r$\;
        }
    }
    $M_l[o] = best\_r$\;
}
\Return{$M_l$}

\caption{Calculate outcome mastery for learner $l$: get best result across items verifying each outcome}
\end{algorithm}


After running the algorithm, each outcome $o$ receives a mastery score based on 
the learner's best result on any verifying item. This choice of using the maximum was deliberately generous. Other aggregation 
rules (such as averages or weighted scores) could also be used if a stricter definition 
is required.

\medskip
From the per–outcome mastery scores $M_l$, we can also define a single 
\textbf{Total Mastery} score for learner~$l$ by summing over all outcomes:
\[
T_l = \sum_{o \in O} M_l[o].
\]
This formulation keeps the distinction between local mastery (per outcome) 
and global mastery (total).

\medskip
Finally, the hierarchical structure of the outcome tree allows for extensions. 
For example, if a node has no direct verifying items, its score can be inferred 
from the mastery of its children. In the same way, priorities or weights can be 
attached to particular skills, so that high-priority areas contribute more strongly 
to the total mastery. These variations show how the same framework can be tailored 
to different learning settings.

\subsection{Case Study Project Integration}

\begin{figure}[h]
    \centering
    \includegraphics[width=0.45\textwidth]{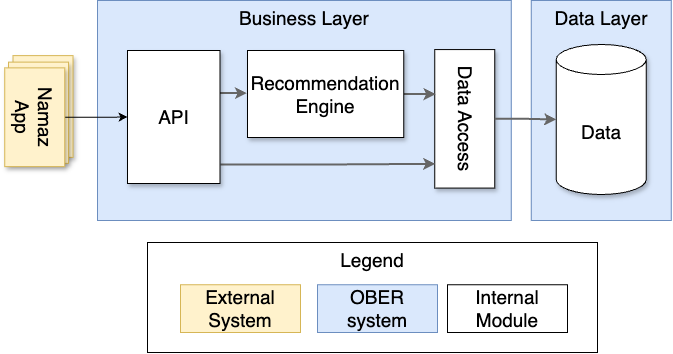}
    \caption{The system architecture}
    \label{fig:architecture}
\end{figure}

Architecturally, OBER is a service consisting of business and data layers (Fig. \ref{fig:architecture}). Following the suggestion in \cite{tan2008learning}, we made the recommendation system into a API service that exist outside of the client application.

\begin{figure}[h]
    \centering
    \includegraphics[width=0.25\textwidth]{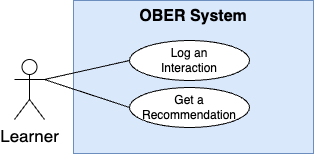}
    \caption{Use-case diagram: System's User-facing Functionality}
    \label{fig:use_case_diagram}
\end{figure}

Fig. \ref{fig:use_case_diagram} demonstrates how simple the client-facing functionality of the system is. Essentially, the learner interacts with items and gets recommendations.

\begin{figure}[htbp]
\centering
\subfloat[Recommender UI\label{fig:recommender_ui}]%
{\includegraphics[width=0.32\linewidth]{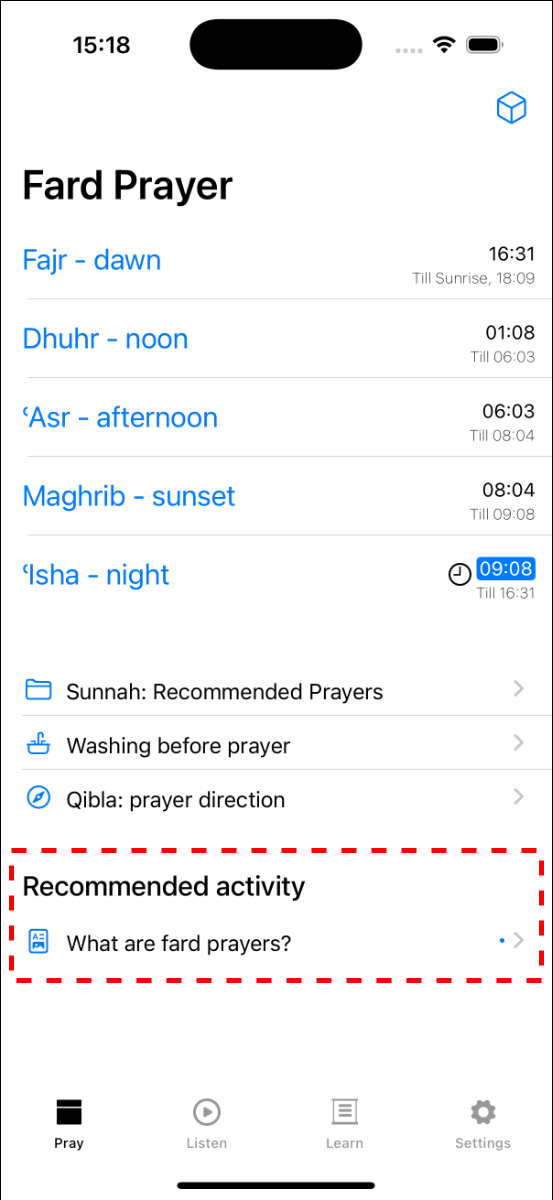}}\hfill
\subfloat[Learning Item\label{fig:learning_item}]%
{\includegraphics[width=0.32\linewidth]{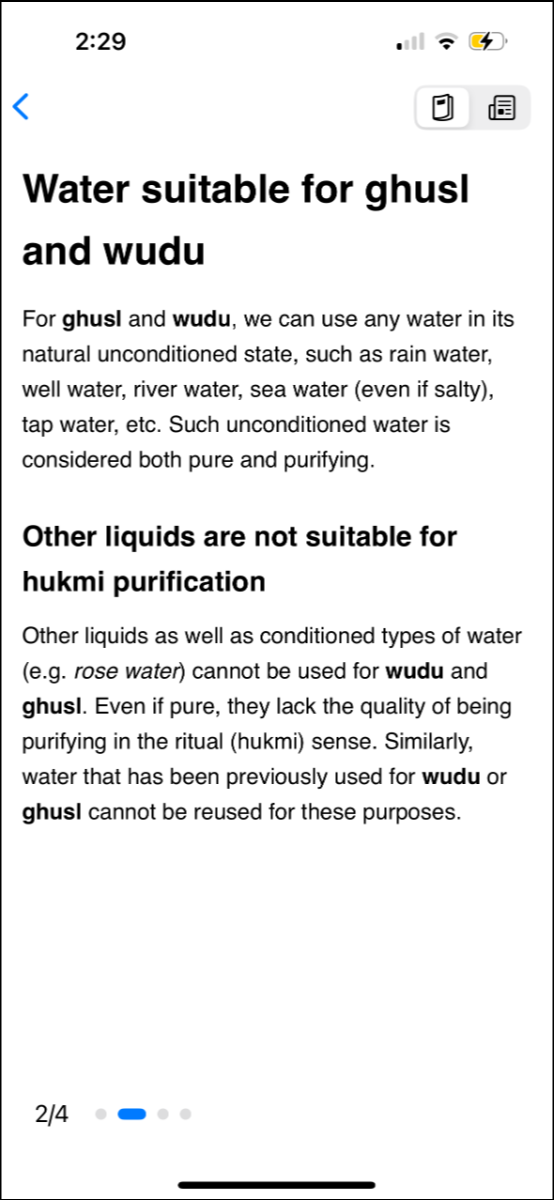}}\hfill
\subfloat[Assessment item\label{fig:assessment_item}]%
{\includegraphics[width=0.32\linewidth]{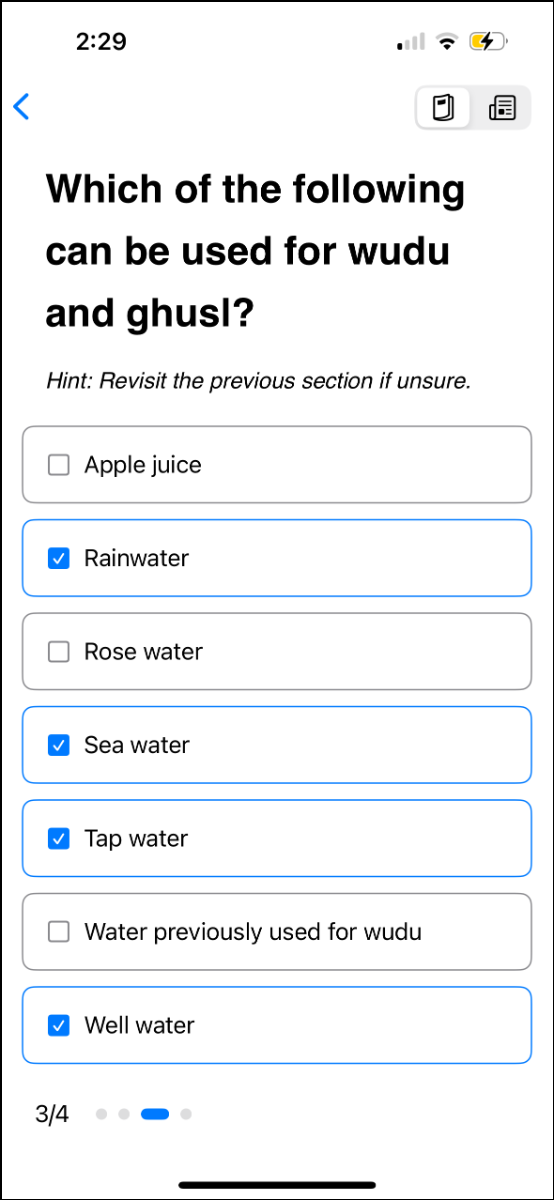}}
\caption{User facing OBER functionality in NamazApp.}
\label{fig:threefigs}
\end{figure}

The recommendations were shown on the main page of NamazApp (Fig. \ref{fig:recommender_ui}). Learning items aimed to deliver the material (Fig. \ref{fig:learning_item}) through text, audio, and images. Then a typical assessment item was served in the form of interactive quizzes where a learner answered various type of questions (Fig. \ref{fig:assessment_item}), including multiple choice, drag-and-drop answers, and more.

As mentioned earlier, an important feature of OBER is its method independence. This is reflected in the architecture, as the recommendation engine is a separate module. The modular design makes it easier to have a variety of recommendation methods. Within NamazApp, we implemented the following methods:

\begin{itemize}
  \item \textbf{Fixed Trajectory:} Follows a predefined standard sequence of items curated by the NamazApp team, same for all learners assigned this method.
  \item \textbf{Collaborative Filtering (CF):} Recommends items based on the preferences of other learners with similar interaction histories.
  \item \textbf{Knowledge‐Based (KB) Filtering:} Suggests items by leveraging \emph{promotes} alignment mappings between items and outcomes.
\end{itemize}


Having alternative recommendation methods allowed us to conduct an experiment to compare their effectiveness from an outcome-based perspective. We describe the evaluation methodology in the following section.




\section{Evaluation}

The main hypothesis of our paper is that OBER and its underlying data schema can enable us to assess the educational effect of recommendations. More specifically, we want to measure which recommendation strategy maximizes the mastery of learning outcomes.

\begin{figure}[h]
    \centering
    \includegraphics[width=0.48\textwidth]{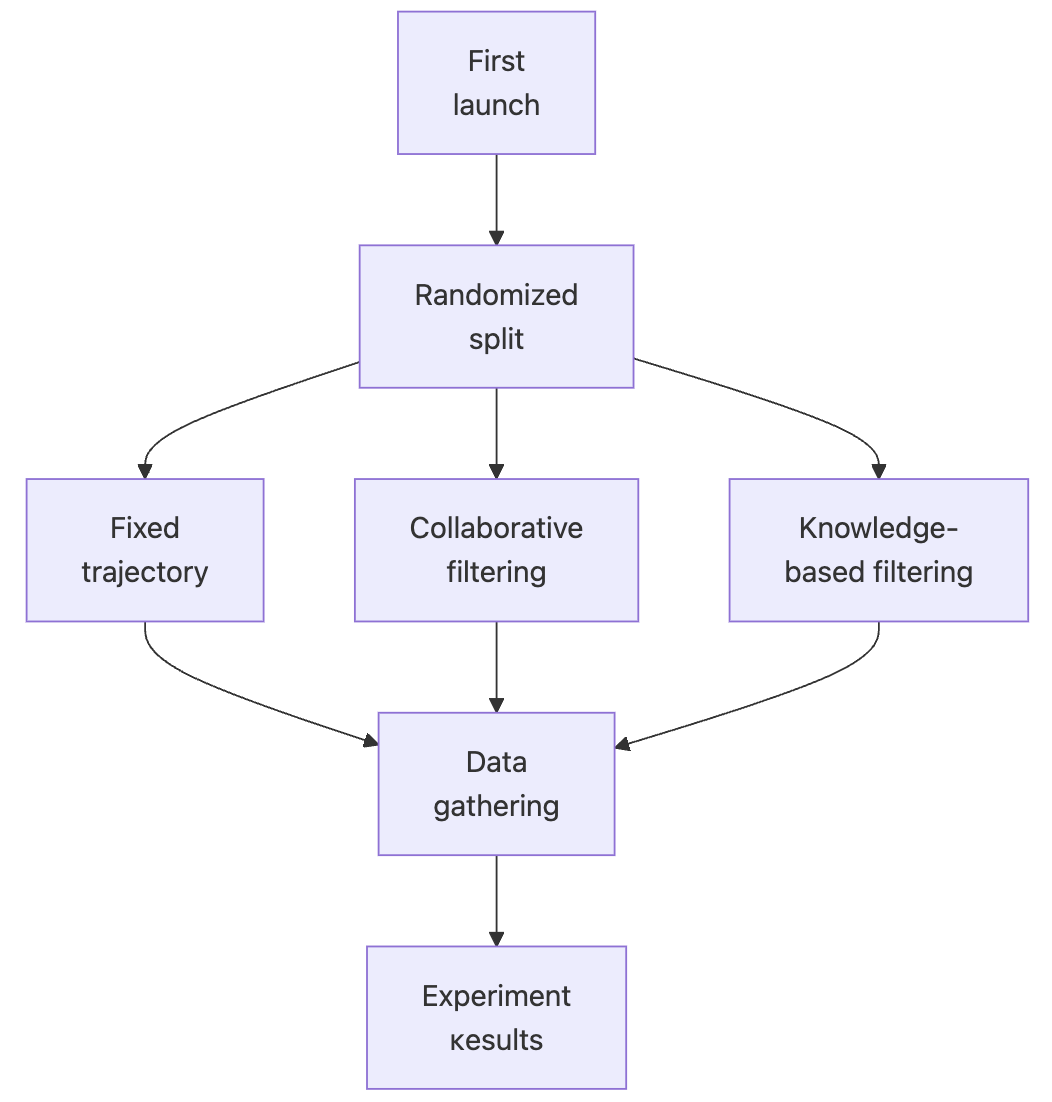}
    \caption{Evaluation flow and stages}
    \label{fig:evaluation_flow}
\end{figure}

To answer the hypothesis, we conducted split-testing (a.k.a. online controlled experiment) \cite{fabijan2017evolution}. As shown in the Fig. \ref{fig:evaluation_flow}, we randomly distributed the learners into three groups. The first group had a fixed predefined set of recommendations, the second group had Collaborative Filtering (CF) as the recommendation algorithm, and the third group received knowledge-based (KB) recommendations. 

The exact choice of recommendation methods was not critical to this study, rather the point was to demonstrate that we can evaluate and compare their educational effectiveness. Still, there are also a few reasons why we chose these three methods. The predefined recommendations set is the baseline designed by NamazApp team. CF is reported as the most popular algorithm in the research on educational recommender system \cite{urdaneta2021recommendation}. KB is also a popular approach that can have many variations \cite{usageContextBased6980102, Semantic7272748, MLAcademia8693719, children9956839}. In our case, it leveraged the \textit{alignment} mappings between \textit{Outcomes} and \text{Items}. At the end, we had a curated list, the most popular recommendation method, and an outcome-based method.

\subsection{Experiment}

The goal of our evaluation is to demonstrate that OBER can (1) seamlessly swap between different recommendation methods, (2) embed outcome-based measurement within normal usage, and (3) compare methods without extra engineering. These translated into following three objectives.

\begin{enumerate}
  \item Method independence: Show that replacing one recommendation method with another requires no changes to logging or mastery calculation.
  \item Embedded measurement: Verify that recommendations can generate complete outcome mastery data as part of natural usage.
  \item Comparability: Confirm that outcome-based metrics (mastery trajectories, time to threshold) are comparable across modules.
\end{enumerate}

Given our objectives, we did not use the common pre-/post-test technique, to show that OBER enables outcome measurement within the recommendation flow without needing separate dedicated testing phases. All mastery data emerged naturally from learner interactions.

We ran the experiment in NamazApp for two weeks for a fraction of users, collecting anonymized logs for slightly over 1800 learners in each group. Then, we analyzed the logs using the logs, outcomes tree and the mastery formula. The exact experiment metrics calculated for each of the three groups are shown in Table \ref{tab:metrics}.

\begin{table}[h]
    \centering
    \renewcommand{\arraystretch}{1.2} 
    \begin{tabular}{|p{1.3cm}|p{5.3cm}|}
        \hline
        \textbf{Metric} & \textbf{Description} \\
        \hline
        Retention & Average number of sessions per learner. \\
        \hline
        Relevance & Average click-through rate for recommendations per learner. \\
        \hline
        Mastery & Average total mastery score per learner. \\
        \hline
    \end{tabular}
    \caption{Experiment metrics used in the study.}
    \label{tab:metrics}
\end{table}

We chose these particular metrics to have a comprehensive view on the effectiveness of each method:

\begin{itemize}
    \item \emph{Retention} is a common business metric, often used as the indicator of how valuable the product is to the end-user.
    \item \emph{Relevance} is the main metric for evaluating recommender systems, which can be measured in several ways (most commonly an explicit or implicit user rating), while we opted for the click-through rate on recommendations.
    \item \emph{Mastery} is the main outcome-based metric for the OBER system and our study overall.
\end{itemize} 

\subsection{Results}

Table~\ref{tab:results_actual} summarizes the key metrics for each recommendation method, computed over the collected sample.

\begin{table}[h]
    \centering
    \begin{tabular}{|l|r|r|r|r|}
        \hline
        \textbf{Method} & \textbf{Learners} & \textbf{Retention} & \textbf{Relevance} & \textbf{Mastery} \\
        \hline
        CF     & 1856     & \textbf{6.16}        & \textbf{0.3033}    & 0.3899   \\
        Fixed  & 1963     & 5.63        & 0.3026    & \textbf{0.4641}   \\
        KB     & 1932     & 5.02        & 0.3007    & 0.4212   \\
        \hline
    \end{tabular}
    \caption{Aggregated experiment metrics by recommendation method.}
    \label{tab:results_actual}
\end{table}

The metrics show that \emph{CF} has the highest retention among the three methods. Yet, the \emph{Fixed} trajectory—while being lower on retention—achieves a higher total mastery. \emph{KB} recommendations perform moderately in mastery while having the lowest retention and relevance.

\begin{figure}[h]
    \centering
    \includegraphics[width=0.95\linewidth]{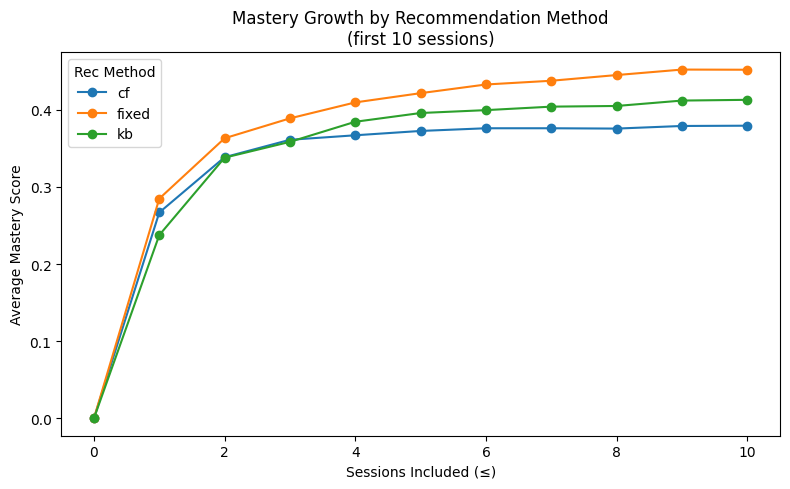}
    \caption{Mastery growth by recommendation method (first ten sessions).}
    \label{fig:mastery_growth}
\end{figure}

Figure~\ref{fig:mastery_growth} illustrates how mastery evolves over the first ten sessions for each method to show a more dynamic view on the static averages above.

From these results, we see that:
\begin{itemize}
  \item \emph{CF} leads on retention but lags in final mastery.
  \item \emph{Fixed} delivers the highest mastery gains over time, with a steeper growth curve than CF and KB.
  \item \emph{KB} falls in between, improving steadily but not matching the fixed trajectory’s peak.
\end{itemize}
Overall, while CF keeps learners coming back, the fixed predefined trajectory yields the strongest learning outcomes.

\section{Discussion}

\subsection{Outcome‐Based Metrics in Non‐Formal Settings}  
Outcome‐based evaluation is central to e‐learning, yet it often goes unmeasured in non‐formal contexts. OBER fills this gap by providing a comprehensive set of metrics—business (retention), recommendation (relevance), and pedagogical (mastery)—that together enable a nuanced assessment of recommendation effectiveness.

\subsection{Empirical Findings}  
Our split‐testing experiment yielded three key observations:
\begin{itemize}
  \item \textbf{Collaborative Filtering (CF)} produced the highest average sessions per learner and a slight lead in CTR, reflecting its emphasis on relevance and engagement through peer‐data.
  \item \textbf{Fixed Predefined Trajectory} achieved the highest total mastery, demonstrating that an expertly curated, non‐personalized sequence can drive superior learning gains.
  \item \textbf{Knowledge‐Based (KB)} filtering—outcome‐oriented and personalized via alignment mappings—delivered mastery results intermediate between CF and Fixed.
\end{itemize}
These results show that optimizing solely for relevance (CF) does not ensure deeper learning. Instead, outcome‐focused approaches—whether curated or alignment‐driven—more reliably support mastery, and a well‐designed fixed trajectory can even outperform a personalized method on learning outcomes.

\subsection{Comprehensive Metrics Enable Decision-making}  
By surfacing both product metrics (retention) and educational metrics (mastery), OBER allows stakeholders to weigh trade‐offs explicitly. Furthermore, analyzing correlations between these dimensions can reveal whether improvements in one metric (e.g.\ CTR) causally influence another (e.g.\ mastery), informing strategic choices (for example, whether boosting engagement is likely to enhance learning).

\subsection{Framework Extensibility}  
OBER’s minimal yet expressive data schema supports easy extension to new domains. For instance, we could:
\begin{itemize}
  \item incorporate learner attributes such as prior knowledge, motivation, or preferences into recommendation logic and mastery calculation;
  \item integrate contextual signals (time of day, location, affective state) to enable context‐aware recommendation strategies.
\end{itemize}

\subsection{Personalizing Outcomes}  
Beyond personalizing item recommendations, OBER can personalize the very outcomes. Learners can be assigned high‐level competencies based on their goals and background, which are refined via outcome hierarchies. Using \emph{verifies}‐type alignment mappings, one can then measure mastery on specific outcomes, enabling evidence‐based, goal‐driven trajectories.

\subsection{Limitations and Future work}  
\begin{itemize}
  \item \textbf{Sample Size and Signifcance:} Our sample of 5.7 thousand data points was sufficient to detect engagement differences. However, a larger experiment with a longer timeframe could allow to discover more subtle correlations between metrics and establish statistically significant differences between recommendation methods.
  \item \textbf{Generalizability:} Our findings are specific to NamazApp: a mobile e-learning system from non-formal educational domain. Validating OBER across other educational contexts is a subject of future work.
\end{itemize}

\section{Conclusion}

We introduced OBER, an outcome‐based educational recommender system, to fill a gap in recommender evaluation: most systems optimize clicks or ratings, not learning.  OBER extends the classic user–item schema with learning outcomes and defines a mastery formula that computes both per‐outcome and total mastery directly from interaction logs.  

We demonstrated OBER’s practicality by integrating it into NamazApp, then running an online controlled experiment comparing three recommendation methods (CF, KB, Fixed).  Our results show that:
\begin{itemize}
  \item CF excels at retention but does not translate clicks into superior mastery.
  \item A fixed, expert‐curated trajectory yields the highest learning gains.
  \item Knowledge‐based alignment methods offer moderate performance.
\end{itemize}

Crucially, all measurements—business (retention), recommendation (CTR), and pedagogical (mastery)—were gathered through the same unified pipeline, without any separate testing phases.  

As further work, we plan to explore hybrid strategies that marry expert curation with adaptive personalization, extend OBER to support context‐aware and goal‐driven recommendations, and validate the framework across diverse educational domains.  By making outcome measurement a first‐class citizen, OBER enables educational recommenders that verifiably increase learning.



\bibliographystyle{ieeetr}
\bibliography{ober}

\begin{thebibliography}{10}

\bibitem{rivera2018recommendation}
A.~C. Rivera, M.~Tapia-Leon, and S.~Lujan-Mora, ``Recommendation systems in education: a systematic mapping study,'' in {\em International Conference on Information Technology \& Systems}, pp.~937--947, Springer, 2018.

\bibitem{drachsler2015panorama}
H.~Drachsler, K.~Verbert, O.~C. Santos, and N.~Manouselis, ``Panorama of recommender systems to support learning,'' in {\em Recommender systems handbook}, pp.~421--451, Springer, 2015.

\bibitem{urdaneta2021recommendation}
M.~C. Urdaneta-Ponte, A.~Mendez-Zorrilla, and I.~Oleagordia-Ruiz, ``Recommendation systems for education: systematic review,'' {\em Electronics}, vol.~10, no.~14, p.~1611, 2021.

\bibitem{spady1994outcome}
W.~G. Spady, {\em Outcome-Based Education: Critical Issues and Answers}.
\newblock American Association of School Administrators, 1994.

\bibitem{europeanQualification}
E.~Union, ``Europass tools: European qualifications framework.'' \url{https://europa.eu/europass/en/europass-tools/european-qualifications-framework}.
\newblock [Accessed: May 15, 2023].

\bibitem{biggs2012student}
J.~Biggs, ``What the student does: Teaching for enhanced learning,'' {\em Higher education research \& development}, vol.~31, no.~1, pp.~39--55, 2012.

\bibitem{askarbekuly2021building}
N.~Askarbekuly, A.~Solovyov, E.~Lukyanchikova, D.~Pimenov, and M.~Mazzara, ``Building an educational product: constructive alignment and requirements engineering,'' in {\em Advances in Artificial Intelligence, Software and Systems Engineering: Proceedings of the AHFE 2021 Virtual Conferences on Human Factors in Software and Systems Engineering, Artificial Intelligence and Social Computing, and Energy, July 25-29, 2021, USA}, pp.~358--365, Springer, 2021.

\bibitem{our_slr}
N.~Askarbekuly and I.~Lukovi{\'c}, ``Learning outcomes, assessment, and evaluation in educational recommender systems: A systematic review,'' {\em arXiv preprint arXiv:2407.09500}, 2024.

\bibitem{burke2002hybrid}
R.~Burke, ``Hybrid recommender systems: Survey and experiments,'' {\em User modeling and user-adapted interaction}, vol.~12, no.~4, pp.~331--370, 2002.

\bibitem{deeplearning2022}
L.~Salau, M.~Hamada, R.~Prasad, M.~Hassan, A.~Mahendran, and Y.~Watanobe, ``State-of-the-art survey on deep learning-based recommender systems for e-learning,'' {\em Applied Sciences}, vol.~12, no.~23, p.~11996, 2022.

\bibitem{reinforcement2022}
M.~M. Afsar, T.~Crump, and B.~Far, ``Reinforcement learning based recommender systems: A survey,'' {\em ACM Computing Surveys}, vol.~55, no.~7, pp.~1--38, 2022.

\bibitem{fuzzyTree7094243}
D.~Wu, J.~Lu, and G.~Zhang, ``A fuzzy tree matching-based personalized e-learning recommender system,'' {\em IEEE Transactions on Fuzzy Systems}, vol.~23, no.~6, pp.~2412--2426, 2015.

\bibitem{blended7104183}
N.~Hoic-Bozic, M.~Holenko~Dlab, and V.~Mornar, ``Recommender system and web 2.0 tools to enhance a blended learning model,'' {\em IEEE Transactions on Education}, vol.~59, no.~1, pp.~39--44, 2016.

\bibitem{ontology10}
J.~Joy, N.~S. Raj, and R.~V.~G., ``Ontology-based e-learning content recommender system for addressing the pure cold-start problem,'' vol.~13, no.~3, 2021.

\bibitem{ersdoDynamicOntology}
M.~Amane, K.~Aissaoui, and M.~Berrada, ``Ersdo: E-learning recommender system based on dynamic ontology,'' vol.~27, p.~7549–7561, jul 2022.

\bibitem{SocialRetrieval2018}
C.~K. Pereira, F.~Campos, V.~Ströele, J.~M.~N. David, and R.~Braga, ``Broad-rsi – educational recommender system using social networks interactions and linked data,'' {\em Journal of Internet Services and Applications}, vol.~9, p.~7, 03 2018.

\bibitem{largescale}
K.~Dahdouh, A.~Dakkak, L.~Oughdir, and A.~Ibriz, ``Large-scale e-learning recommender system based on spark and hadoop,'' {\em Journal of Big Data}, vol.~6, no.~1, p.~2, 2019.

\bibitem{elearning_keys}
T.~C. Reeves, ``Keys to successful e-learning: Outcomes, assessment and evaluation,'' {\em Educational Technology}, vol.~42, no.~6, pp.~23--29, 2002.

\bibitem{askarbekuly2020combining}
N.~Askarbekuly, A.~Sadovykh, and M.~Mazzara, ``Combining two modelling approaches: Gqm and kaos in an open source project,'' in {\em Open Source Systems: 16th IFIP WG 2.13 International Conference, OSS 2020, Innopolis, Russia, May 12--14, 2020, Proceedings 16}, pp.~106--119, Springer, 2020.

\bibitem{usercentric7994718}
S.~Fazeli, H.~Drachsler, M.~Bitter-Rijpkema, F.~Brouns, W.~v.~d. Vegt, and P.~B. Sloep, ``User-centric evaluation of recommender systems in social learning platforms: Accuracy is just the tip of the iceberg,'' {\em IEEE Transactions on Learning Technologies}, vol.~11, no.~3, pp.~294--306, 2018.

\bibitem{scratch8651403}
J.~Cárdenas-Cobo, A.~Puris, P.~Novoa-Hernández, J.~A. Galindo, and D.~Benavides, ``Recommender systems and scratch: An integrated approach for enhancing computer programming learning,'' {\em IEEE Transactions on Learning Technologies}, vol.~13, no.~2, pp.~387--403, 2020.

\bibitem{Askarbekuly2024}
N.~Askarbekuly and N.~Aničić, ``{LLM examiner: automating assessment in informal self-directed e-learning using ChatGPT},'' {\em Knowledge and Information Systems}, vol.~66, no.~10, pp.~6133--6150, 2024.

\bibitem{AlShurunbulali2010}
A.~H. ibn 'Ammar~al Shurunbulali, {\em {Ascent to Felicity: A Manual on Islamic Creed and Hanafi Jurisprudence}}.
\newblock California: White Thread Press, 2010.
\newblock Paperback edition.

\bibitem{tan2008learning}
H.~Tan, J.~Guo, and Y.~Li, ``E-learning recommendation system,'' in {\em 2008 International conference on computer science and software engineering}, vol.~5, pp.~430--433, IEEE, 2008.

\bibitem{fabijan2017evolution}
A.~Fabijan, P.~Dmitriev, H.~H. Olsson, and J.~Bosch, ``The evolution of continuous experimentation in software product development: from data to a data-driven organization at scale,'' in {\em 2017 IEEE/ACM 39th International Conference on Software Engineering (ICSE)}, pp.~770--780, IEEE, 2017.

\bibitem{usageContextBased6980102}
K.~Niemann and M.~Wolpers, ``Creating usage context-based object similarities to boost recommender systems in technology enhanced learning,'' {\em IEEE Transactions on Learning Technologies}, vol.~8, no.~3, pp.~274--285, 2015.

\bibitem{Semantic7272748}
P.~Montuschi, F.~Lamberti, V.~Gatteschi, and C.~Demartini, ``A semantic recommender system for adaptive learning,'' {\em IT Professional}, vol.~17, no.~5, pp.~50--58, 2015.

\bibitem{MLAcademia8693719}
H.~Samin and T.~Azim, ``Knowledge based recommender system for academia using machine learning: A case study on higher education landscape of pakistan,'' {\em IEEE Access}, vol.~7, pp.~67081--67093, 2019.

\bibitem{children9956839}
M.~Moradi, K.~R. Fard, and M.~Y. Akhlaqi, ``A recommender system method for children’s education using mobile technology,'' {\em IEEE Access}, vol.~10, pp.~123679--123696, 2022.

\end{thebibliography}





\end{document}